\documentclass[letterpaper, 10 pt, conference]{ieeeconf}  

\IEEEoverridecommandlockouts                              

\usepackage[dvipdfmx]{graphicx} 
\usepackage{multirow}
\usepackage{arydshln}
\usepackage{url}
\usepackage{spverbatim}
\usepackage{ascmac}
\usepackage{algorithm}
\usepackage{overpic}
\usepackage[noend]{algorithmic}
\usepackage{amsmath}

\usepackage{tcolorbox}
\tcbuselibrary{breakable} 

\title{\LARGE \bf
Compact Task-Aligned Imitation Learning for Laboratory Automation
}

 \author{
 Kanata Suzuki$^{1}$,
 Hanon Nakamura$^{2}$,
 Kana Miyamoto$^{2}$,
 and Tetsuya Ogata$^{2,3}$
 \thanks{
 $^{1}$Kanata Suzuki is affiliated with Spatial Robotics Research Center, Fujitsu Limited., Kanagawa 211-8588, Japan. 
 $^{2}$Hanon Nakamura, Kana Miyamoto, and Tetsuya Ogata are affiliated with Faculty of Science and Engineering, Waseda University, Tokyo 169-8050, Japan. 
 $^{3}$Tetsuya Ogata is also at the National Institute of Advanced Industrial Science and Technology, Tokyo 100-8921, Japan. 
 E-mail:{\tt\small suzuki.kanata@fujitsu.com}
 }}

\begin{document}

\maketitle
\thispagestyle{empty}
\pagestyle{empty}


\begin{abstract}

Robotic laboratory automation has traditionally relied on carefully engineered motion pipelines and task-specific hardware interfaces, resulting in high design cost and limited flexibility. 
While recent imitation learning techniques can generate general robot behaviors, their large model sizes often require high-performance computational resources, limiting applicability in practical laboratory environments.
In this study, we propose a compact imitation learning framework for laboratory automation using small foundation models. 
The proposed method, TVF-DiT, aligns a self-supervised vision foundation model with a vision-language model through a compact adapter, and integrates them with a Diffusion Transformer-based action expert. 
The entire model consists of fewer than 500M parameters, enabling inference on low-VRAM GPUs.
Experiments on three real-world laboratory tasks—test tube cleaning, test tube arrangement, and powder transfer—demonstrate an average success rate of 86.6\%, significantly outperforming alternative lightweight baselines.
Furthermore, detailed task prompts improve vision-language alignment and task performance. 
These results indicate that small foundation models, when properly aligned and integrated with diffusion-based policy learning, can effectively support practical laboratory automation with limited computational resources.

\end{abstract}

\section{INTRODUCTION}
\label{sec1}

Recent advances in artificial intelligence have significantly accelerated and enhanced research activities across scientific domains. 
However, many laboratory operations still rely heavily on manual procedures, creating a strong demand for robotic laboratory automation~\cite{king2009automation}. 
Research in laboratory automation can be broadly categorized into two directions: automation of experimental procedures through hardware-integrated robotic systems~\cite{steiner2019organic}\cite{burger2020mobile}\cite{szymanski2023autonomous}, and optimization of experimental condition exploration~\cite{macleod2020self}\cite{granda2018controlling}\cite{coley2020autonomous}\cite{kanda2022robotic}\cite{hase2018phoenics}.

Conventional approaches typically design robot motions in advance and automate entire workflows by controlling dedicated interfaces of laboratory equipment.
While such systems enable highly precise and repeatable execution of experiments, they incur substantial design costs and are difficult to apply to auxiliary tasks that are not central to the experimental objective (e.g., organizing or cleaning laboratory tools). 
In other words, existing autonomous laboratory systems primarily emphasize closed-loop experimental optimization, whereas lightweight learning-based robotic manipulation for everyday laboratory tasks remains underexplored.

One promising approach to address this limitation is imitation learning~\cite{Suzuki2023}. 
Recent Vision Language Action (VLA) models~\cite{kim2024openvla}\cite{rt-x}\cite{pi05} have demonstrated that general robotic behaviors can be learned from demonstration data. 
However, these models typically rely on large-scale models (LLMs~\cite{grattafiori2024llama} or VLMs~\cite{wu2025qwen}\cite{beyer2024paligemma}) as backbones, resulting in large model sizes that require high-performance computing resources for local deployment. 
Such requirements are not well-suited for laboratory automation scenarios, where learning robot tasks themselves is not the primary research objective.

\begin{figure}[tb]
    \centering
    \includegraphics[width=\columnwidth]{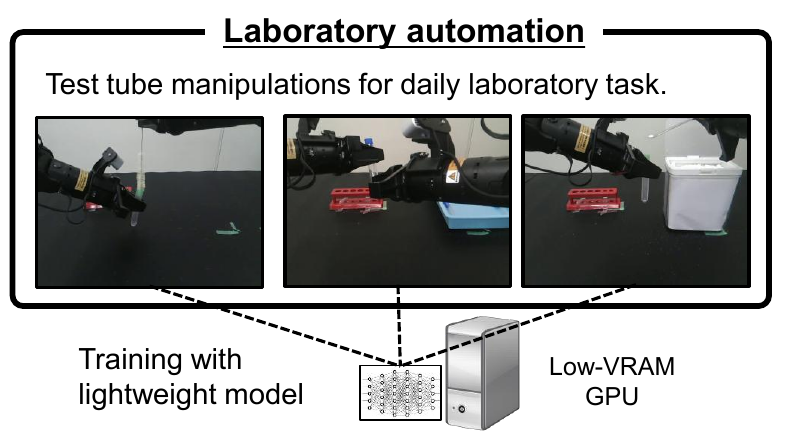}
    \caption{
Overview of this study for a laboratory automation framework with a compact imitation learning model. 
We focus on robotic test tube manipulation under limited computational resources.
    }
    \label{fig1}
\end{figure}


To address this challenge, we propose a compact imitation learning framework for laboratory automation. 
By leveraging relatively compact self-supervised vision foundation models and vision-language foundation models as backbones, our approach enables both training and inference on GPUs with limited VRAM. 
Furthermore, we demonstrate that aligning the vision foundation model with task-specific prompts facilitates more efficient learning.
As a representative case of laboratory automation, we focus on robotic test tube manipulation tasks and validate the effectiveness of the proposed method (Fig.~\ref{fig1}). 
Test tube operations are routinely performed in chemical laboratories and include cleaning, powder handling, and arrangement tasks. 
Although these actions are intuitive for humans, they require precise alignment and continuous motion generation for robots, making it difficult to manually predefine trajectories.

\section{RELATED WORK}
\label{sec2}

\subsection{Laboratory Automation with Real Robot Operation}

Robotic operations in laboratory automation have predominantly relied on approaches that pre-design robot motions as well as dedicated instruments and system architectures~\cite{steiner2019organic}\cite{burger2020mobile}\cite{szymanski2023autonomous}\cite{sasaki2025robotic}\cite{roch2018chemos}\cite{christensen2021data}.
Burger et al. developed an automated experimental system using a mobile robot, and it autonomously executed 688 experiments in eight days~\cite{szymanski2023autonomous}.
Sasaki et al. proposed a chemical experiment automation system integrating a jig interface, motion demonstration interface, jig control interface, and a mobile manipulator~\cite{sasaki2025robotic}. 
While these approaches enable precise and repeatable experimental execution, they remain difficult to apply to auxiliary tasks that are not central to the experimental objective. 
In practical laboratory environments, the experimental workflow includes various routine tasks such as preparation and cleanup, which are typically performed by students or technical staff. 
Automating these everyday tasks is essential for advancing the overall goal of laboratory autonomy.


Several studies have explored learning-based approaches for optimizing robotic experimental operations~\cite{pizzuto2024accelerating}\cite{kadokawa2023learning}\cite{yamaguchi2014learning}\cite{wang2025pipetting}. 
Pizzuto et al. formulated a sample scraping task in chemical experiments as a model-free reinforcement learning problem within a task-specific simulation environment~\cite{pizzuto2024accelerating}.
Kadokawa et al. formulated powder weighing as a reinforcement learning problem and proposed a recurrent neural network-based policy that balances conservative and proactive behaviors~\cite{kadokawa2023learning}. 
They further validated their approach on real hardware via sim-to-real transfer using domain randomization techniques. 
Although these studies demonstrate the effectiveness of learning-based methods, they generally require dedicated simulation environments~\cite{schenck2017reasoning}, which may limit their applicability in practical laboratory settings.

\subsection{Imitation Learning}
Imitation learning has emerged as a promising approach for achieving general robotic behaviors~\cite{chi2024diffusionpolicy}\cite{act}\cite{Toyoda2022}\cite{Suzuki2024}. 
Among these approaches, VLA models construct general action policies by leveraging a VLM pretrained on large-scale offline data as a backbone, followed by task-specific fine-tuning through imitation learning using demonstration data~\cite{kim2024openvla}\cite{rt-x}\cite{pi05}. 
These models have demonstrated the ability to generate generalizable robotic behaviors across tasks. 
However, standard VLMs are typically trained with textual supervision~\cite{wu2025qwen}, which strengthens image-language alignment but may reduce sensitivity to fine-grained geometric details. 
VLA models compensate for this limitation through large-scale robot demonstration datasets; nevertheless, their performance in out-of-distribution environments such as laboratory settings is not guaranteed.

Recent research has focused on extending VLA models by incorporating video generation models~\cite{liu2026world}\cite{kim2026cosmos} or reward prediction mechanisms~\cite{pi05}. 
For example, COSMOS Policy~\cite{kim2026cosmos} integrates a generative video model that predicts future observation frames.
The model optimizes actions within the latent space and enables task learning with limited demonstrations. 
Similarly, $\pi_{0.5}$~\cite{pi05} augments VLA architectures with reward prediction and feedback mechanisms. 
By introducing an additional VLM to evaluate action quality and reinforcing training with reward signals, they improve real-world task success rates. 
While these approaches enhance the representational capacity of VLAs, they tend to increase overall model size. 
Although model compression techniques such as quantization~\cite{xu2026qvla} have been explored, most VLA models remain at the billion-parameter scale.


Motivated by these studies, our work adopts a different approach.
We combine a self-supervised vision foundation model with a vision-language foundation model, and learn a lightweight adapter to connect their representations.
Image foundation models, which also serve as backbones for video generation models, have demonstrated human-level or even superhuman performance in various recognition tasks.
In laboratory automation, however, deploying large-scale generalist models is often impractical.
Instead, specializing smaller models for specific tasks can be more efficient and reliable.
Our method leverages geometrically consistent patch-token representations obtained through self-supervised learning.
These representations are efficiently integrated with patch-token representations from a vision-language foundation model using only a small number of additional parameters.
As a result, the proposed approach enables efficient action learning and inference under limited VRAM constraints.

\section{PROPOSED METHOD}
\label{sec3}

\begin{figure*}[tb]
    \centering
    \includegraphics[width=2.0\columnwidth]{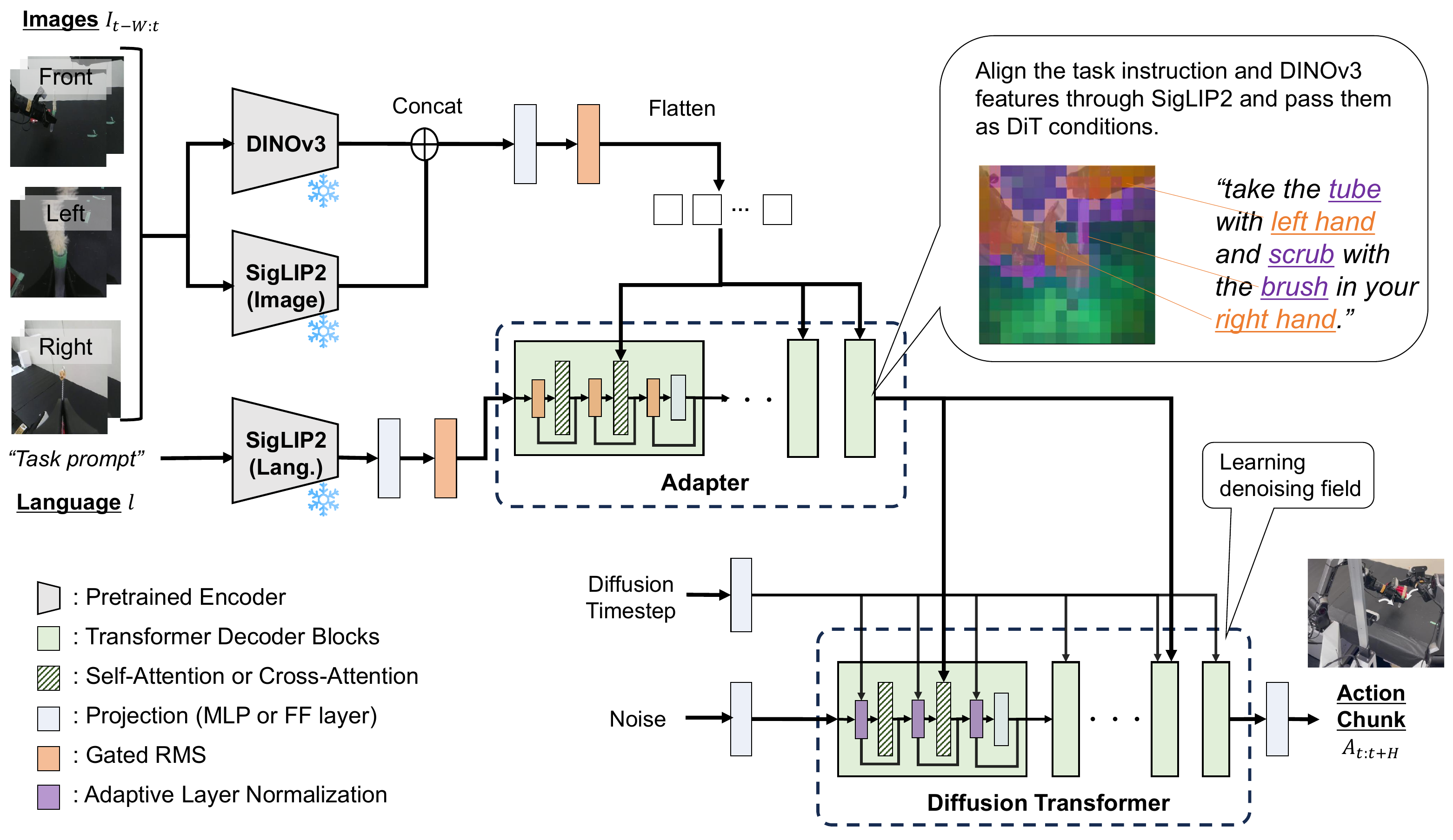}
    \caption{Architecture of the proposed TVF-DiT framework. DINOv3 and SigLIP2 extract geometric and language-aligned representations, which are fused via a lightweight Adapter. The resulting task-conditioned tokens are used as cross-attention keys and values in a Diffusion Transformer that predicts action chunks through conditional flow matching.}
    \label{fig2}
\end{figure*}

We propose an imitation learning framework that combines a Task-aligned Vision Foundation model with a Diffusion Transformer (TVF-DiT). 
An overview of the proposed method is shown in Fig.~\ref{fig2}. 
The model aligns generic object-centric geometric representations extracted by a self-distillation-based vision foundation model with task prompts through a shared feature-mapping vision-language model. 
The detailed parameter configuration of the proposed method is summarized in Table I. 
A notable characteristic of our approach is that it is constructed entirely from pretrained models with fewer than 500M parameters in total.

\subsection{Vision and Language Encoder}
\label{sec3_1}
\subsubsection{Vision Encoder: }
RGB images $I_t$ captured from cameras mounted on the robot are first processed by the vision encoder, where $t$ denotes the timestep. 
We use three cameras in total: one mounted at the front of the robot and the other two attached to each end-effector, i.e., $I_t=[i_t^1,i_t^2,i_t^3]$.. 
As vision encoders, we employ DINOv3~\cite{dino} and SigLIP2~\cite{siglip2}. 
Specifically, we use the pretrained models “vit\_small\_patch16\_dinov3.lvd1689m” and “siglip2-base-patch16-224”, respectively, and keep their weights frozen during the training phase. 
DINOv3 is a self-distillation-based vision foundation model that learns invariance to viewpoint and scale changes through data augmentation. 
It is therefore expected to capture fine-grained shape information and local structural features required for laboratory automation tasks. 
In this work, we deliberately adopt relatively small vision backbone models due to inference hardware constraints. 
Each image is encoded into patch tokens of identical spatial resolution by the respective encoders. 
And then, the patch tokens from DINOv3 and those from the SigLIP2 vision encoder are concatenated along the feature dimension. 
This design facilitates subsequent alignment with task prompts, as described later.

\begin{table}[t]
    \centering
    \begin{tabular}{ccll}
        \multicolumn{4}{c}{TABLE I: Structure of Proposed Method} \\
        \hline
        & & Model Type/Layers & Parameter \\
        \hline \hline
        \multirow{5}{*}{VL Encoder} & DINOv3~\cite{dino} 
        & vit\_small, patch16 
        & 21 M \\ 
        \cdashline{2-4}
        & SigLIP2~\cite{siglip2}
        & vit\_base, patch16
        & 375 M \\
        \cdashline{2-4}
        & \multirow{3}{*}{Adapter} 
          & Projection $\times$ 2 & \multirow{3}{*}{33 M} \\ 
        & & GatedRMS~\cite{qiu2026unified} $\times$ 2 & \\ 
        & & Trans. Dec. Block $\times$ 8 & \\ 
        \hline
        \multirow{2}{*}{Action Expert} & \multirow{2}{*}{DiT} 
          & Projection $\times$ 3 & \multirow{2}{*}{45 M} \\ 
        & & Trans. Dec. Block $\times$ 8 & \\ 
        \hline
    \end{tabular}
    \begin{flushleft}
        * All layers have leaky GeLU activation. \\
        ** All Transformer Block have 2048 dim for FF layer and 8 heads. \\
        *** Projection dimensions are 512. \\
    \end{flushleft}
\end{table}

\subsubsection{Language Encoder: }
The language input $l$ (task prompt) is tokenized using the SigLIP2 language encoder~\cite{siglip2}. 
To align language tokens with visual features, the language embeddings are fed into a learnable Adapter module. 
In this process, the encoded image tokens are provided as keys and values to the Transformer Decoder blocks within the Adapter. 
All modality tokens are projected into a shared embedding dimension. 
To prevent weight outliers caused by training with limited demonstration data, we apply GatedRMS normalization~\cite{qiu2026unified}. 
This design choice was adopted based on empirical observations that it stabilizes training. 
SigLIP2 is a contrastive VLM trained with a sigmoid-based binary classification objective, projecting both language and visual representations into a shared embedding space~\cite{siglip2}. 
Consequently, image regions that are semantically aligned with the prompt become strongly correlated through inner-product computations in the cross-attention layers of the Transformer Decoder. 
These correlations are further enhanced by the object-centric geometric representations extracted by DINOv3 and concatenated at the patch-token level. 
By providing appropriately designed task instructions as prompts, the model can achieve stronger alignment with the intended task. 
In the experiments, we evaluate how including explicit descriptions of the robot arms and manipulated objects in the prompts affects task performance.

\subsection{Action Expert}
The Action Expert, which predicts robot actions, is built upon the Diffusion Transformer (DiT~\cite{chi2024diffusionpolicy}).
The model primarily consists of multiple Adaptive Layer Normalization (AdaLN~\cite{peebles2023scalable}) functions and cross-attention layers.
Task-conditioned tokens extracted from the observation $o_t=[I_{t-W},\dots,I_{t},l]$ by the vision-language encoder
are used as keys and values in the cross-attention layers of the DiT. 
Here, $W$ denotes the number of past observation frames; in this work, we set $W = 2$.
Cross-attention layers are inserted once every two Transformer blocks within the DiT. 
In the proposed model, we avoid excessive conditioning via self-attention in order to reduce computational cost while maintaining adaptive behavior generation in response to environmental changes.

The DiT takes random noise as input and outputs an action sequence (action chunk) $A=[a_t,a_{t+1},\dots,a_{t+H-1}]$ where $H$ denotes the prediction horizon and is set to $H = 32$ in this study. 
Offline imitation learning is performed using Conditional Flow Matching (CFM)~\cite{cfm-1}\cite{cfm-2}. 
The model parameters are optimized by minimizing the following loss function:
\begin{equation}
L^{\tau}(\theta)
=
E_{p(A_t^,o_t),q(A_t^\tau|A_t)}||v_\theta(A_t^\tau,o_t)-u(A_t^\tau|A_t)||^2
\end{equation}
where subscripts denote robot timesteps and superscripts denote flow-matching timesteps, with $\tau \in [0,1]$.
By minimizing this loss, the model $v_\theta$ learns a denoising vector field $u$ that transforms random noise into the desired action sequence.
Following the previous study~\cite{pi05}, the flow-matching timestep $\tau$ is sampled from a beta distribution and injected as a conditioning signal into the AdaLN functions.
During inference, we perform 10 denoising steps and execute model prediction asynchronously with the robot control loop. 
If the model prediction is not completed within a control cycle, we apply actions from the previously predicted action chunk to maintain control stability.
The final action command is computed as a weighted average of the predicted action chunk~\cite{act}.

\begin{figure}[tb]
    \centering
    \includegraphics[width=\columnwidth]{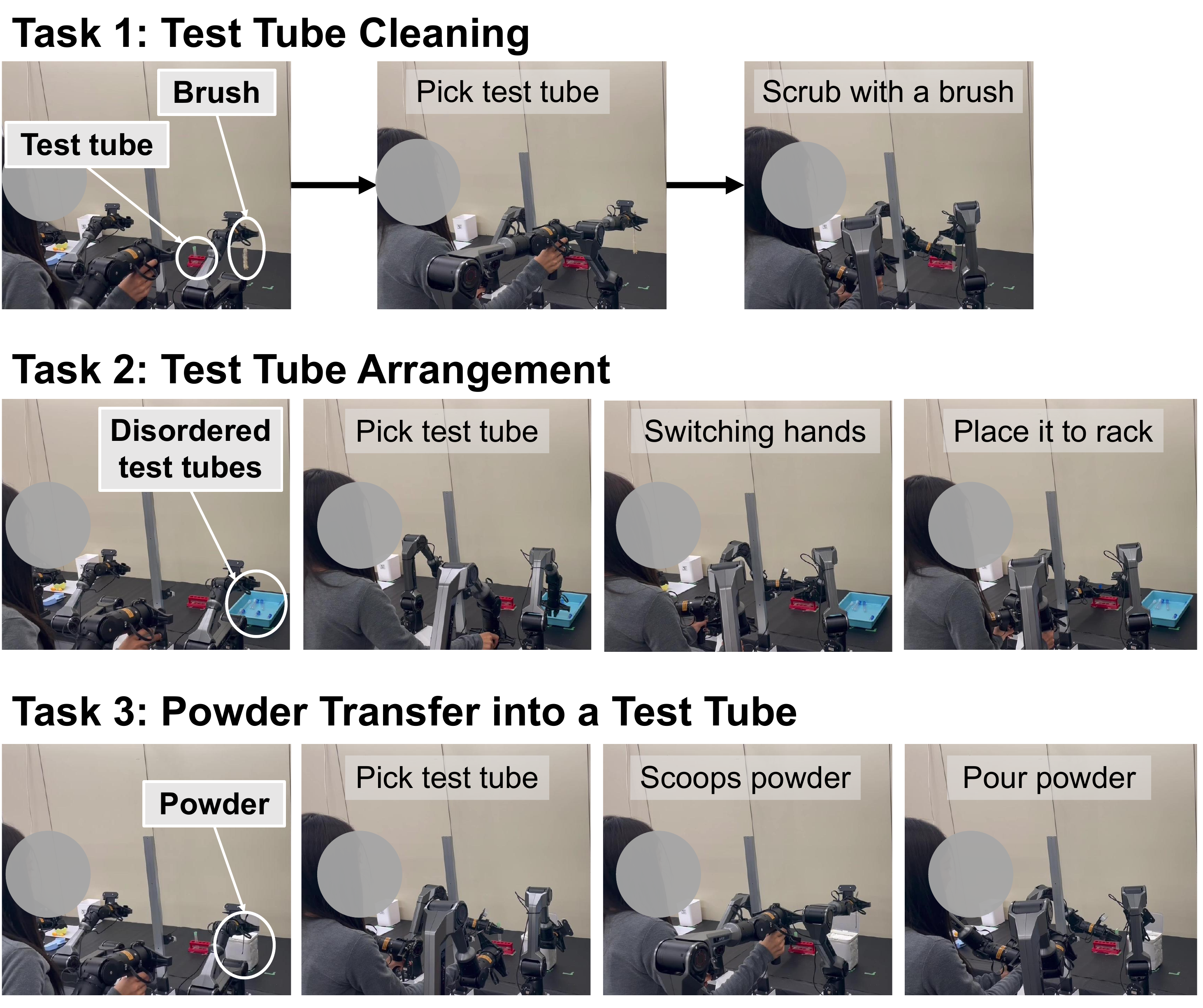}
    \caption{
Task 1: Test tube cleaning requiring precise insertion and continuous scrubbing motion. 
Task 2: Test tube arrangement requiring collision avoidance and bimanual coordination. 
Task 3: Powder transfer requiring sequential scooping and pouring manipulation. 
These tasks evaluate geometric precision, object selection, and continuous action generation.
    }
    \label{fig3}
\end{figure}

\section{EXPERIMENTS}
\label{sec4}

\subsection{Robot Task}
To evaluate the effectiveness of the proposed method, we conducted learning experiments using a dual-arm mobile manipulator, CobotMagic~\cite{cobotmagic}. 
We designed three experimental tasks, described below.

\textbf{Task 1: Test Tube Cleaning}
In this task, the robot performs a coordinated dual-arm cleaning operation on a test tube. The left arm first grasps a test tube placed in a rack. The robot then inserts the brush into the narrow interior of the tube and executes a continuous scrubbing motion along the inner surface.
This task requires precise geometric alignment between the brush and the tube opening, stable bimanual coordination during insertion, and sustained periodic motion generation for scrubbing. The limited clearance inside the tube makes the task highly sensitive to positional errors, and visual feedback is essential for maintaining alignment during motion.

\textbf{Task 2: Test Tube Arrangement}
In this task, multiple test tubes are randomly scattered on a tray. The robot must select a target tube, grasp it without colliding with surrounding objects, and place it into a designated rack location.
This task involves object selection under cluttered conditions, collision-aware motion generation, and coordinated handover between arms during placement. The spatial relationships between objects vary across trials, requiring adaptive behavior rather than replaying fixed trajectories.

\textbf{Task 3: Powder Transfer into a Test Tube}
In the powder transfer task, the robot scoops granular powder using a spoon and pours it into a test tube held by the opposite arm. The task requires sequential transitions between scooping, transporting, and pouring.
Unlike rigid-object manipulation, powder handling involves partially deformable and dynamically changing states. The robot must maintain appropriate spoon orientation to prevent spillage and generate smooth motion during pouring.

An overview of the tasks is shown in Fig.~\ref{fig3}. 
Tasks 1 and 3 were selected as laboratory automation tasks involving deformable object manipulation~\cite{Fujii2022sii}\cite{saito2025learning}, while Task 2 was designed based on interviews with researchers who conduct chemical experiments, targeting a routine laboratory activity that is often considered tedious and prone to being neglected. 
Each task involves different manipulation elements, such as grasping, alignment, and continuous motion generation. 
The goal is to evaluate whether a vision-based action generation method can generalize across diverse laboratory operations.

\subsection{Training}
Demonstration data for imitation learning were collected via teleoperation using a leader–follower control setup. 
For the test tube cleaning task, 500 episodes were collected, while 400 episodes were collected for each of the other tasks, all recorded at 50 Hz. 
The total recording time amounted to approximately eight hours. 
As observation inputs to the model, we used RGB images of size $224 \times 224 \times 3$. 
As action outputs, the model predicts 14 DoF motor values (joint angles and grippers) corresponding to both robot arms.

Training was performed as offline imitation learning. 
To normalize scale differences, all motor values were normalized to the range $[-1, 1]$ using soft min-max normalization. 
We used a batch size of 16 and a gradient accumulation frequency of 8. 
The exponential moving average (EMA) decay rate was set to 0.999, and gradient clipping was applied with a threshold of 1.0. 
The model was optimized using AdamW with a learning rate of $1 \times 10^{-4}$ and weight decay of $1 \times 10^{-8}$. 
Training was conducted for 400,000 iterations. 
All training was performed on a single desktop machine equipped with an NVIDIA GeForce RTX 4090 (24GB VRAM), and the total training time was approximately 18 hours.

\subsection{Evaluation}
Inference was performed on an NVIDIA GeForce RTX 4060 (8GB VRAM) on the robot's control PC, and the robot control frequency was set to 50 Hz. 
The proposed model can be executed on an onboard PC with a low-VRAM GPU. 
For real-robot experiments, we conducted the following evaluations for each task.

\subsubsection{Comparison of Model Architectures}

We evaluated the effect of model architecture on task success rates. 
We considered two alternative configurations similar to the proposed architecture: (i) a standalone VLM model, and (ii) a combination of a vision foundation model and a LLM. 
For each setting, pretrained models were selected so that the total number of parameters was comparable to that of the proposed method. 
We compared these configurations with the proposed method based on task success rates.

\subsubsection{Comparison of Training Prompt}

As discussed in the previous section, the conditioned tokens extracted by the vision-language encoder are expected to depend on the granularity of the prompts used during training. 
To evaluate this, we prepared three types of prompts for each task: (a) a task-irrelevant prompt, (b) a concise prompt describing the task, and (c) a detailed prompt describing the task. 
We compared performance in terms of task success rate. 
The key difference among these prompts lies in the level of task description, particularly whether they include explicit references to objects manipulated during the task.

\section{RESULTS AND DISCUSSION}
\label{sec5}

\subsection{Comparison of Model Architecture}

Table II presents the task success rates for different vision-language encoder configurations. 
The first row shows the results of simple playback, while the second and third rows correspond to (i) a standalone VLM and (ii) a combination of a vision foundation model and a LLM, respectively. 
As the baseline VLM, we used SmolVLM2-256M~\cite{marafioti2025smolvlm}. 
SmolVLM2 is a lightweight VLM designed for resource-efficient settings and is commonly adopted in compact VLA systems. 
It represents a compact multimodal foundation model capable of operating under limited computational resources and serves as a substitute for the entire encoder in the proposed framework. 
Similarly, as the baseline language model, we employed SmolLM2-135M~\cite{allal2025smollm2}. SmolLM2 is a lightweight decoder-only LLM with compact Transformer blocks, frequently used as a backbone for VLMs and robotics applications. 
In our experiments, we replaced SigLIP2 with SmolLM2 in the proposed architecture. 
For all learning-based methods except playback, we used the detailed task prompt (c) during the training phase. 
The details of the training prompts are described in the following subsection.


The results show that simple playback failed in all trials, indicating the inherent difficulty of the tasks. 
Methods (i) and (ii) achieved overall success rates of 20.0\% and 36.6\%, respectively. In contrast, the proposed method (iii) achieved the highest performance, with an overall success rate of approximately 86.6\%. 
These results suggest that the self-distillation-based vision foundation model, DINOv3, effectively captures object-level geometric features, and that the use of SigLIP2 as a shared feature-mapping vision-language model enables effective alignment between task prompts and visual observations.
 In contrast, methods (i) and (ii) did not achieve sufficient alignment between language and vision. 
This indicates that multimodal learning typically requires large-scale data to achieve strong cross-modal alignment.
It is consistent with prior findings on robot foundation models~\cite{pi05}.

\begin{table}[t]
    \centering
    \begin{tabular}{l|ccc|c}
        \multicolumn{5}{c}{TABLE I: Success rates in test tube manipulations} \\
        \hline
                              & Task 1 & Task 2 & Task 3 & Ave.  \\
        \hline \hline
        Playback & 0/10 & 0/10 & 0/10 & 0.0\% \\
        \hline
        (i) w/ VLM  & \multirow{2}{*}{1/10} & \multirow{2}{*}{2/10} & \multirow{2}{*}{3/10} & \multirow{2}{*}{20.0\%} \\
        (SmolVLM2) & & & & \\
        \hline
        (ii) w/ VF \& LLM & \multirow{2}{*}{2/10} & \multirow{2}{*}{4/10} & \multirow{2}{*}{5/10} & \multirow{2}{*}{36.6\%} \\
        (DINOv3 \& SmolLM2) & & & & \\
        \hline
        (iii) TVF-DiT & \multirow{2}{*}{8/10} & \multirow{2}{*}{9/10} & \multirow{2}{*}{9/10} & \multirow{2}{*}{86.6\%} \\
        (DINOv3 \& SigLIP2) & & & & \\
        \hline
    \end{tabular}
\end{table}

\begin{figure*}[tb]
    \centering
    \includegraphics[width=2.0\columnwidth]{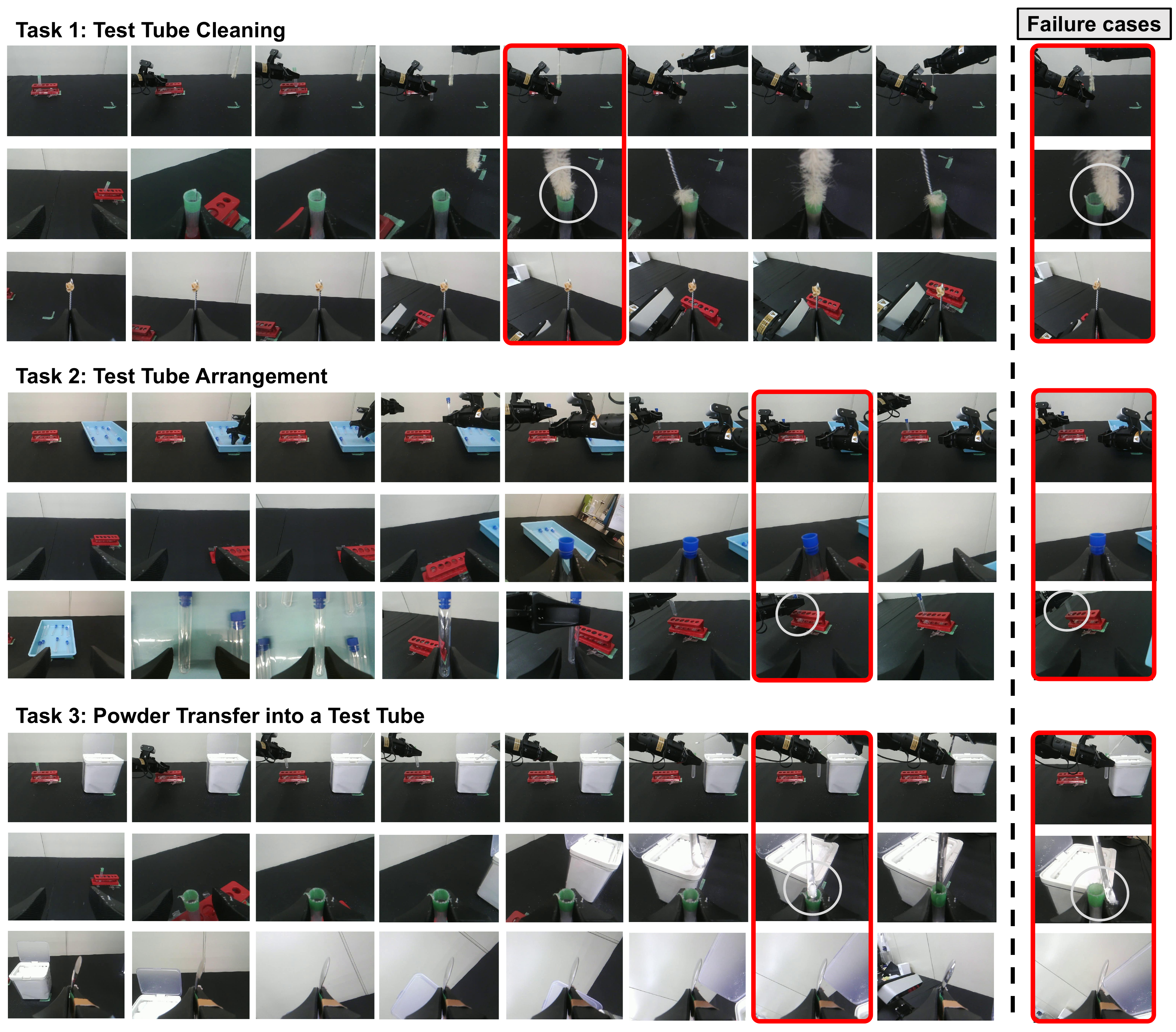}
    \caption{Representative execution sequences generated by the proposed method. Successful trials demonstrate continuous and coordinated manipulation across all tasks. Failure cases (highlighted) mainly occur during fine alignment or placement, indicating sensitivity to small geometric errors.}
    \label{fig5}
\end{figure*}

\begin{figure}[tb]
    \centering
    \includegraphics[width=1.0\columnwidth]{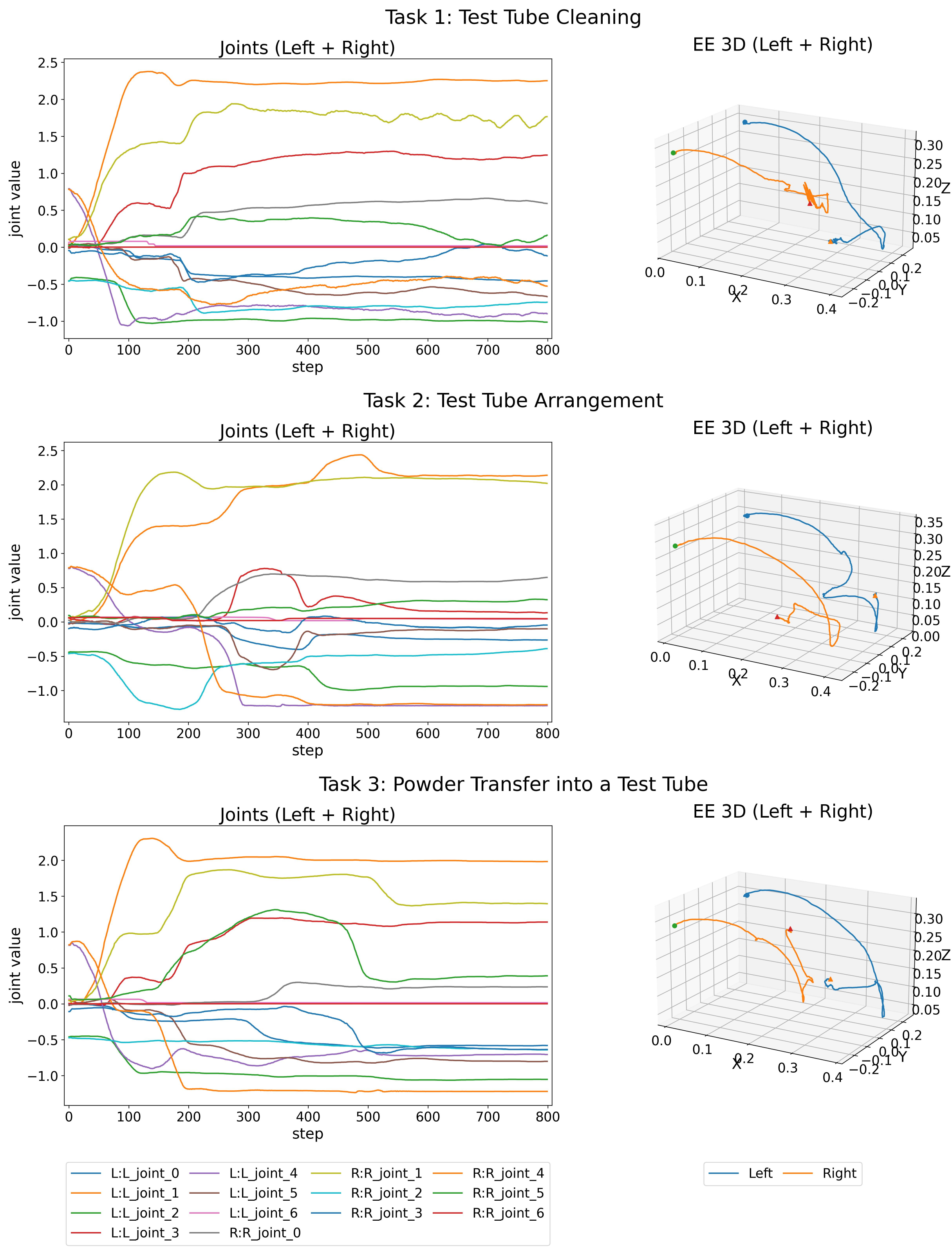}
    \caption{Joint trajectories and 3D end-effector (EE) paths during inference. 
    Consistent EE paths across episodes demonstrate robustness of the learned policy.}
    \label{fig4}
\end{figure}

Examples of generated motions for each task are shown in Fig.~\ref{fig4} and Fig.~\ref{fig5}. 
In the test tube cleaning task (Task 1), successful trials demonstrated continuous insertion and scrubbing motions of the brush inside the tube. 
Failure cases involved premature termination of the scrubbing motion or failure to insert the brush to test tube. 
In the test tube arrangement task (Task 2), the robot successfully avoided collisions with surrounding tubes while grasping the target tube and placing it at the designated location. 
Most failure cases occurred during placement, where slight positional misalignment prevented correct insertion into the rack. 
In the powder transfer task (Task 3), the robot consistently scooped powder and transferred it into the test tube. 
Visual inspection confirmed that the amount of powder transferred was stable across most trials.
In all tasks, the generated trajectories exhibit smooth and coordinated motion patterns, indicating stable diffusion-based action prediction (Fig.~\ref{fig5}).

\begin{figure*}[tb]
    \centering
    \includegraphics[width=2.0\columnwidth]{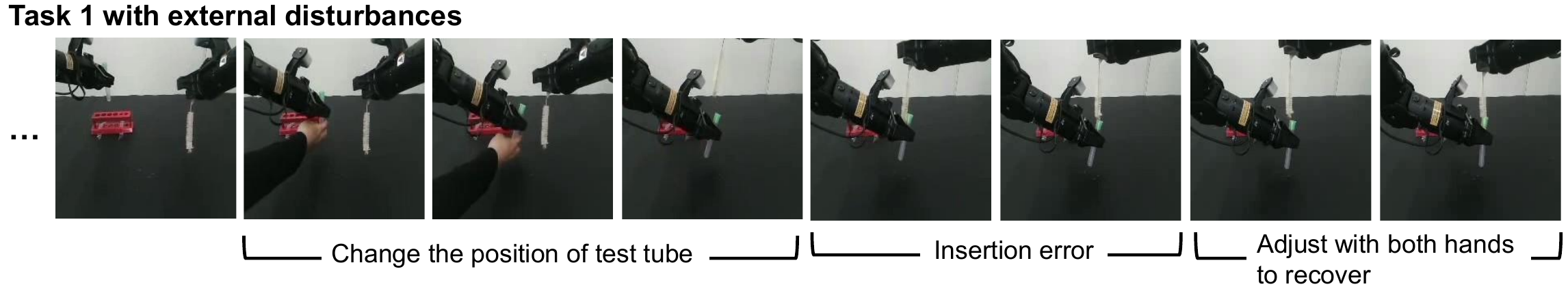}
    \caption{Recovery behavior under external disturbance during test tube cleaning. When the tube orientation is manually perturbed, the model re-generates actions based on updated visual observations and resumes the task. This demonstrates the adaptive capability of the diffusion-based policy.}
    \label{fig6}
\end{figure*}

We further evaluated robustness under external disturbances in the test tube cleaning task, as shown in Fig.~\ref{fig6}. When the orientation of the grasped test tube was manually perturbed during execution, the robot regenerated actions based on visual feedback and successfully resumed the cleaning motion.
These results indicate that the proposed method exhibits a certain degree of adaptability under dynamic environmental changes.

\subsection{Performance Evaluation with Trained Prompts}
Table III presents task success rates under different prompts used in the training phase. 
All results in this table were obtained using the proposed TVF-DiT model.
The first row corresponds to (a) a task-irrelevant prompt, the second to (b) a concise task description, and the third to (c) a detailed task description.
For prompt (a), we used:
"this is a test sentence that is not related to the robot's task."
For prompts (b) and (c), we used: "brush the tube." / "take the tube with left hand and scrub with the brush in your right hand." for Task 1, "set the tube." / "pick up the tube placed in the box with right hand, receive it with left hand, and place it in the tube rack." for Task 2, and "put powder into a tube." / "take the tube with left hand and scoop up the powder with right hand and pour it into the tube." for Task 3.

The results show that task-irrelevant prompts significantly degraded performance, indicating insufficient alignment between language and visual information.
Concise prompts also resulted in reduced performance for all tasks except Task 2. We attribute this to whether the prompt explicitly mentions the objects manipulated during the task. 
In Task 2, the robot interacts with only one type of object (test tubes), allowing the concise prompt to sufficiently guide attention.
In contrast, detailed prompts achieved the highest success rates across all tasks. These findings demonstrate that carefully designed prompts can substantially improve task performance by enhancing cross-modal alignment within the Adapter.

Interestingly, models trained with detailed prompts (c) exhibited reduced performance when evaluated using the irrelevant prompt (a), but maintained performance when evaluated using the concise prompt (b).
This suggests that only a subset of patch tokens is critical for task execution; however, richer supervision during training—through more detailed prompts—contributes to improved alignment and overall learning performance.

\begin{table}[t]
    \centering
    \begin{tabular}{l|ccc|c}
        \multicolumn{5}{c}{TABLE II: Success rates in trained prompts with TVF-DiT} \\
        \hline
                              & Task 1 & Task 2 & Task 3 & Ave.  \\
        \hline \hline
        (a) Task-irrelevant prompt  & \multirow{2}{*}{0/10} & \multirow{2}{*}{0/10} & \multirow{2}{*}{0/10} & \multirow{2}{*}{0.0\%} \\
        (13 word) & & & & \\
        \hline
        (b) Concise task prompt & \multirow{2}{*}{2/10} & \multirow{2}{*}{9/10} & \multirow{2}{*}{5/10} & \multirow{2}{*}{53.3\%} \\
        (3--5 words) & & & & \\
        \hline
        (c) Detailed task prompt & \multirow{2}{*}{8/10} & \multirow{2}{*}{9/10} & \multirow{2}{*}{9/10} & \multirow{2}{*}{86.6\%} \\
        (15--23 words) & & & & \\
        \hline
    \end{tabular}
\end{table}

\subsection{Limitations}
First, the models used in this study are lightweight. In particular, we adopted the smallest publicly available DINOv3 variant. Using larger vision foundation models may enable more precise geometric representation and potentially improve task success rates.
Although the proposed framework aims to enable laboratory automation under limited VRAM constraints, future improvements in GPU hardware may allow integration of larger backbone models. Incorporating model compression techniques such as quantization~\cite{xu2026qvla} to support larger models under constrained resources is an important direction for future work.

Second, the tasks evaluated in this study represent only a subset of laboratory automation scenarios. Expanding the task domain is necessary for broader applicability.
For tasks that require higher precision, performance may depend heavily on the amount of demonstration data. Improving data collection strategies or integrating reward feedback mechanisms could further enhance performance.
Finally, fully automating laboratory workflows would require extending the system to include navigation and broader robotic capabilities, which remains an important avenue for future research.

\section{CONCLUSION}
\label{sec6}

This paper presented TVF-DiT, a compact imitation learning framework for laboratory automation that integrates self-supervised visual representation learning, vision-language alignment, and diffusion-based action generation. 
By combining DINOv3 and SigLIP2 through a compact adapter and employing a Diffusion Transformer as the action expert, the proposed model maintains fewer than 500M parameters while achieving strong real-robot performance.

Experimental results on three test tube manipulation tasks demonstrated that TVF-DiT significantly outperforms alternative lightweight model configurations. 
The findings highlight the importance of combining geometrically consistent visual features with task-aligned language conditioning. 
Moreover, the study showed that detailed task prompts enhance multimodal alignment and improve execution success rates. 

Although this study focuses on specific tube manipulation tasks, the approach provides a scalable direction toward practical laboratory automation with limited computational resources. 
Future work includes expanding the task domain, improving data collection efficiency, and exploring model compression techniques such as quantization to further enhance deployability.



 \section*{ACKNOWLEDGMENT}
 This work was supported by JST PRESTO, Japan, Grant Number JPMJPR24T4.

\bibliographystyle{IEEEtran}
\bibliography{main}

\end{document}